\documentclass{IEEEcsmag}

\usepackage[colorlinks,urlcolor=blue,linkcolor=blue,citecolor=blue]{hyperref}
\expandafter\def\expandafter\UrlBreaks\expandafter{\UrlBreaks\do\/\do\*\do\-\do\~\do\'\do\"\do\-}
\usepackage{upmath,color}
\usepackage{booktabs} 
\usepackage{xcolor}
\usepackage[normalem]{ulem}

\jmonth{September/October}
\jname{IEEE Internet Computing}
\jtitle{IEEE Internet Computing}
\pubyear{2026}

\begin{document}

\sptitle{Special Issue on Wearable Computing}
\title{Affective Agent: On-Device Personalized Intervention Reasoning for Wearable Systems}

\author{Reina Mun}
\affil{Harvard University, Cambridge, MA, USA}
\author{Zishen Wan}
\affil{Harvard University, Cambridge, MA, USA}
\author{Vijay Janapa Reddi}
\affil{Harvard University, Cambridge, MA, USA}

\markboth{Special Issue on Wearable Computing}{AFFECTIVE AGENT}

\begin{abstract} 
Affective computing has advanced wearable state inference, but on-device reasoning about whether, when, and how to intervene remains challenging. We present \emph{Affective Agent}, a three-layer reference architecture for personalized intervention reasoning under uncertainty on wearable-class hardware. It combines a compact sub-billion-parameter language model with physiological evidence, context, and user history to decide whether, when, and how to intervene, without cloud dependency or per-user retraining. The architecture is organized into three interacting layers (perception, personalization, and reasoning), adapting to individual users through host-managed structured memory evolution rather than per-user weight updates. We instantiate Affective Agent in indoor environmental quality control and evaluate it on held-out, simulator-generated longitudinal scenarios spanning physiological variation, context, signal quality, and intervention history. Results show that memory-driven personalization and two-pass structured reasoning improve intervention decisions within this synthetic evaluation. By moving the decision layer on-device, this work demonstrates a path from wearable state inference toward closed-loop, personalized intervention on wearable-class hardware.

\end{abstract}

\maketitle
\begingroup
\renewcommand{\thefootnote}{}
\footnotetext{%
\tiny
\copyright~2026 IEEE. Personal use of this material is permitted.
Permission from IEEE must be obtained for all other uses, in any current or future
media, including reprinting/republishing this material for advertising or promotional
purposes, creating new collective works, for resale or redistribution to servers or
lists, or reuse of any copyrighted component of this work in other works.
Digital Object Identifier 10.1109/MIC.2026.3732091}
\endgroup
\section{Introduction}
\label{sec:intro}
\begin{figure*}[h!]
\centering
\includegraphics[width=\textwidth]{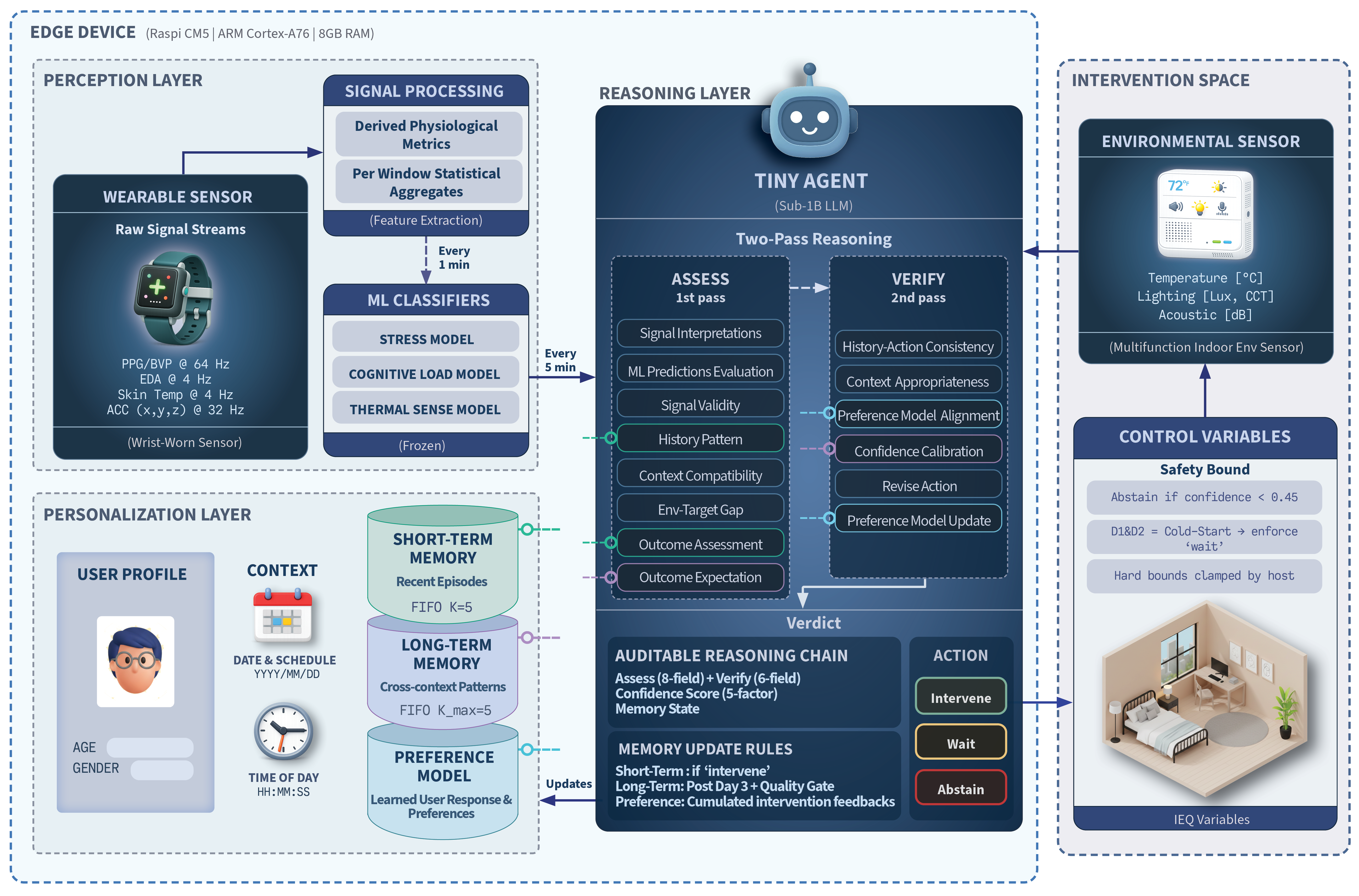}
\caption{\textbf{The Affective Agent Architecture.} (Left) A perception layer converts wearable physiological signals into structured evidence. (Center) A sub-1B reasoning layer produces a bounded intervention decision: intervene, wait, or abstain. (Bottom left) A personalization layer maintains user-specific episodic and semantic memory that conditions each decision. After each cycle, the outcome updates the personalization layer, closing a fully on-device loop. No data leaves the device.}
\label{fig:overview}
\end{figure*}

\chapteri{A}ffective computing concerns systems that sense, interpret, and respond to human affective and psychophysiological states. In wearable systems, recent work has made physiological state inference increasingly feasible: wrist-worn sensors can estimate stress from signals including electrodermal activity (EDA), blood volume pulse (BVP), and heart rate variability (HRV)~\cite{Abdalrazaq2024StressStudents,Zhu2023WristEDA}. However, translating these inferences into timely and meaningful personalized interventions remains challenging. Existing adaptive intervention systems, including JITAIs, can tailor support to changing user states, but persistent challenges remain in adaptivity, receptivity, contextual sensing, and decision rules~\cite{vanGenugten2025,NahumShani2026JITAI}. A wearable may detect elevated heart rate, but that signal alone cannot determine whether the user is exercising, stressed about a deadline, or simply drinking coffee, nor what response is appropriate given the user's current context and prior history. This motivates a decision layer that can reason jointly over physiological evidence, context, and individual history when determining whether, when, and how to intervene.

Consider a software engineer in her room equipped with smart lighting and climate control whose wrist-worn sensor detects rising stress during a focused afternoon coding session. A simple intervention policy might respond with a behavioral prompt, such as a breathing reminder or a self-regulation nudge, even when that prompt is poorly timed and may interrupt concentration. This example illustrates why an inferred physiological state alone is insufficient to determine the timing and form of an intervention.

This challenge is compounded by two practical constraints. First, intervention policies may rely on pre-specified decision rules that do not adapt across users or over time~\cite{vanGenugten2025}. Second, personalization remains difficult in practice because the same physiological signal can call for different responses across individuals and contexts, while privacy, latency, and power constraints can make continual off-device reasoning or per-user retraining undesirable. As compact language models become available at sub-billion parameter scales, building on advances in TinyML and ultra-low-power inference, there is both an opportunity and a need for principled architectures.

We address this through \emph{Affective Agent}, an on-device architecture designed around five properties: (1) on-device reasoning under uncertainty, (2) persistent user-specific memory, (3) personalization through state evolution rather than model retraining, (4) bounded and interpretable decision outputs, and (5) privacy-preserving local operation. The architecture is organized around three layers: perception, personalization, and reasoning (Figure~\ref{fig:overview}), using a compact sub-1B language model with structured memory. 

To ground this framework, we instantiate the architecture in occupant-centric indoor environmental quality (IEQ) control as one exemplar of a wearable intervention space, including temperature, lighting, and acoustics. Compared with JITAIs that deliver behavioral prompts, IEQ interventions can operate ambiently without requiring active user engagement, sustained attention, or an immediate behavioral response. Their effects may also be gradual and produce delayed or noisy feedback. Appropriate actuation therefore depends on physiology, activity, task demands, environmental state, prior outcomes, and learned user preferences.

\textbf{Contributions.} Affective Agent achieves the above properties through three design contributions:
\begin{enumerate}
  \item[{\ieeeguilsinglright}] \textbf{Memory-based personalization.} We show that on-device structured memory and a staged preference life-cycle improve user-specific intervention routing without per-user retraining.
  \item[{\ieeeguilsinglright}] \textbf{Three-layer reference architecture.} We present a design that separates physiological perception, personalization, and intervention reasoning into three interacting layers, enabling privacy-preserving decision-making under tight compute constraints.
  \item[{\ieeeguilsinglright}] \textbf{Structured supervision for a tiny agent.} We demonstrate that a compact sub-1B model can perform structured intervention reasoning when trained with synthetic scenario generation, a reasoning-complexity curriculum, and explicit two-pass reasoning designed for limited parametric capacity.
\end{enumerate}

While we evaluate IEQ as one concrete intervention domain, the proposed architecture is intended as a reusable framework for on-device intervention reasoning in other wearable settings, from consumer wearables such as Samsung Galaxy Watch and Fitbit devices that already surface stress-related measures to emerging health wearables for elder care and cognitive-load management. 

As compact models become increasingly feasible on resource-constrained edge hardware, the question shifts from \emph{can we sense?} to \emph{can we reason about what to do?} Affective Agent examines this question through an on-device reference architecture.


\section{Related Work}
\label{sec:related}

The Affective Agent architecture draws on three bodies of work: wearable intelligence for well-being intervention, personalization methods under resource constraints, and agentic reasoning on edge hardware.

\subsection{Wearable Intelligence}
\label{sec:wearable_intelligence}
For wearable systems to support well-being, they must go beyond sensing to decide \emph{when} and \emph{how} to intervene. Robust pipelines now support stress, thermal comfort inference from wrist-based signals~\cite{Abdalrazaq2024StressStudents,Zhu2023WristEDA}, and hybrid edge-cloud architectures make on-device intelligence increasingly plausible~\cite{Elfouly2025WearableSurvey}, yet many existing systems rely on cloud-side inference or rule-based decision logic.

Just-in-time adaptive interventions (JITAIs) target this gap by delivering support at the moment a person is most receptive. Reviews identify limited adaptivity and receptivity as persistent limitations~\cite{vanGenugten2025,NahumShani2026JITAI}, but on-device reasoning that integrates physiological evidence, user history, and context remains underexplored. Our work recasts this missing decision layer as a structured on-device reasoning problem.

\subsection{Personalization Methods}
\label{sec:personalization_methods}
Affective intervention requires personalization: the same physiological signal may call for different responses depending on the individual and context. Subject-specific models substantially outperform generalized baselines in wearable emotion recognition~\cite{Li2024PersonalizedEmotion}. Longitudinal affect prediction improves when sensor streams are combined with contextual information~\cite{Yang2024AffectForecasting}. Yet personalization in deployed systems often remains coarse-grained~\cite{Matthews2025PersonalisationDMH}. The deeper challenge is achieving meaningful personalization without per-user retraining under wearable constraints. Affective Agent addresses this by decoupling personalization from model weights entirely, relying instead on structured on-device memory and a staged preference life-cycle.

\subsection{Agentic Reasoning on Edge Hardware}
\label{sec:edge_reasoning}
Language models have enabled agentic systems capable of reasoning and decision-making under uncertainty, but these capabilities have been largely realized through cloud-hosted models. Advances in TinyML and on-device inference are beginning to change this. Recent work has demonstrated function-calling with small language models deployed at the edge
(e.g., TinyAgent)~\cite{Yin2024TinyAgent}, and multi-agent coordination on mobile devices (e.g., CAMPHOR)~\cite{Liu2024CAMPHOR}.

However, these systems focus on tool routing and API execution rather than memory-conditioned intervention reasoning. Bridging this gap requires bounded decision-making under resource constraints and noisy physiological observations. The Affective Agent architecture bridges this gap, coupling a compact on-device model with structured memory to perform context-sensitive intervention reasoning under physiological uncertainty.

\section{The Affective Agent Architecture}
\label{sec:architecture}

As compact language models become increasingly deployable on resource-constrained edge hardware, the wearable computing community faces a design question that goes beyond sensing and classification. Building a wearable that senses physiological state is increasingly well established. Building one that reasons about what to do with what it senses, moving from on-device state inference to on-device intervention reasoning, requires a principled architecture. This section presents the Affective Agent architecture, designed to satisfy the five properties introduced in Section~\ref{sec:intro}.

Affective Agent addresses this through three interacting layers that form a closed loop on-device (Figure~\ref{fig:overview}). The perception and reasoning layers address property~(1); the personalization layer addresses properties~(2) and~(3); the structured output schema addresses property~(4); and the fully on-device design addresses property~(5). 

The perception layer converts raw sensor streams into compact physiological evidence. The personalization layer maintains user-specific memory that accumulates across sessions. The reasoning layer, powered by a compact sub-1B language model, combines current evidence with memory to produce a bounded intervention decision: intervene, wait, or abstain. After each decision cycle, the outcome feeds back to the personalization layer's memory to inform future cycles. No data leaves the device.

\subsection{Perception Layer}
\label{sec:perception}
The perception layer (Figure~\ref{fig:overview}, top left) converts raw multimodal physiological signals into a compact, structured evidence representation that a sub-1B model can reason over within a single decision cycle.

Raw sensor streams are not passed directly to the language model. Instead, the host (the runtime on the device that orchestrates sensing, memory, and model invocation) performs local feature extraction per 5-minute window. For each window, it computes statistical aggregates along with signal quality indicators so the model knows when to trust the signal.

The compact evidence representation is then passed to the reasoning layer together with outputs from three frozen task-specific classifiers: a 3-class XGBoost model for stress prediction (baseline, stress, amusement), a Random Forest for cognitive load versus no load, and a binary KNN classifier for thermally comfortable versus uncomfortable states. 

We treat these classifier outputs as auxiliary evidence rather than ground truth. The perception layer reports what it observes but does not decide whether to act on that evidence. That judgment belongs to the reasoning layer, which weighs the evidence against context, memory, and environmental state.

\subsection{Personalization Layer}
\label{sec:personalization}
The personalization layer is the system's persistent user-state layer (Figure~\ref{fig:overview}, bottom left). It maintains a bounded, continuously updated representation of the user (schedule, short- and long-term memory, and a preference model), which conditions how the reasoning layer interprets evidence and selects actions. 

The same physiological pattern may call for different responses. What is an effective intervention for one individual may be ineffective for another, and what works at one point in time may not hold as context or accumulated preference evidence shifts. Without a mechanism to carry this individual history forward, every decision would be made from population-level defaults regardless of what the system has previously observed about the user.

Personalization is implemented through host-managed structured memory that carries user-specific state between the host and the agent. At each decision step, the host retrieves the current memory state and passes it to the agent. The agent then reasons over this memory alongside current evidence and determines whether to update memory. The host validates this proposed intent against explicit write-governance rules before applying it. The language model itself remains stateless across calls.

\subsubsection{Episodic and Semantic Memory}
Because intervention decisions depend on both what happened recently and what the system has learned over time, we maintain two structurally distinct memory channels, borrowing terminology from cognitive science: short-term \emph{episodic} memory (most recent events) and long-term \emph{semantic} memory (generalized knowledge). 

Short-term episodic memory preserves recent intervention episodes. It gives the agent a compact account of what has recently been tried and whether those actions appeared helpful, so the next decision can reflect immediate history rather than treat each cycle as independent. To respect prompt-size constraints, a bounded FIFO buffer (capacity of five) stores what the agent has tried, the context, and the observed near-term physiological response. 

Long-term semantic memory stores learned generalizations derived from accumulated observations over time. The system might learn that a particular user's physiological baseline is lower during morning sessions, or that thermal adjustments have been consistently effective under cognitive load. These generalizations are more durable and slower-moving than episodic memory, allowing the agent to situate current observations within a learned model of the user rather than relying only on recent history.

\subsubsection{Preference Model Life-cycle}
The preference life-cycle governs how strongly user-specific evidence is weighted relative to population defaults. Three stages mark this maturity: \emph{population default} (first two calendar days, cold-start regime), \emph{provisional} (emerging individual patterns begin to influence decisions), and \emph{learned} (confirmed preferences take precedence). A preference state advances to learned only after at least 20 feedbacks and at least 5 days of activity, preventing premature commitment to sparse early evidence. Learned patterns can be revised when new evidence conflicts with them, allowing the system to track gradual preference drift over time.

Memory persists locally as compact per-user JSON. Episodic entries store intervention modality and value, context, timestamp, and measured effectiveness. Semantic memory uses bounded learned-pattern and context-summary stores, with only selected entries inserted into each prompt. This bounds prompt-visible memory while the language model remains stateless across calls.

\subsection{Reasoning Layer}
\label{sec:reasoning}

The reasoning layer (Figure~\ref{fig:overview}, center) is where the system decides what to do. It integrates current physiological evidence, persistent user-specific state from the personalization layer, and situational constraints to determine whether the appropriate response is to intervene, wait, or abstain. 

We implement the layer as a fine-tuned sub-1B language model that operates over a fixed structured schema. Rather than receiving an open-ended prompt, the model takes a predefined input schema and must return a schema-constrained decision contract with bounded action and reasoning fields.

At every decision window, the agent receives the current 5-minute physiological aggregates, outputs from the frozen classifiers, visible context, current environmental conditions, and bounded summaries of short-term and long-term memory. 

\begin{figure}[!h]
\centering
\includegraphics[width=\columnwidth]{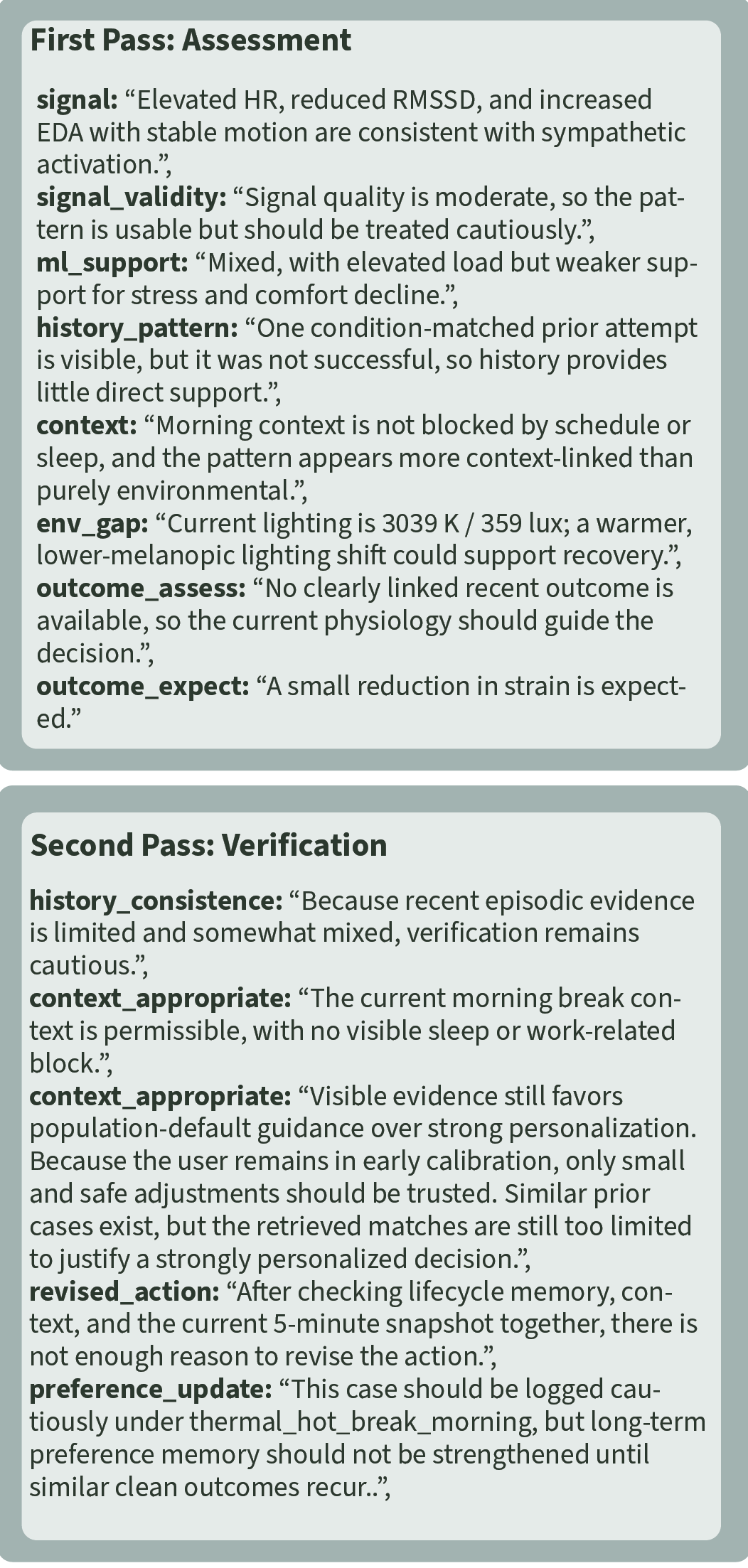}
\caption{\textbf{Two-Pass Reasoning Example.} (Top) First pass. (Bottom) Second pass.}
\label{fig:two_pass}
\end{figure}

A central challenge for sub-1B models is that they cannot perform reliable multi-step intervention reasoning when all required checks are left implicit. To address this, we structure the reasoning process in two explicit passes (Figure~\ref{fig:two_pass}):

The first pass organizes the available evidence into explicit assessment fields: physiological interpretation (sympathetic activation consistent with stress), signal quality (moderate, usable but cautious), classifier support (mixed), history pattern (one prior attempt, unsuccessful), context fit (morning break, not blocked by schedule), environmental mismatch (current lighting suggests a warmer, lower-CCT adjustment could help), and anticipated effect.

The second pass verifies and revises that judgment by checking history consistency, context appropriateness, preference alignment, and confidence calibration. For the engineer, this might mean recognizing that her preference model is still in the provisional stage, so the system tempers personalization and applies only safe, small adjustments. This forces the model to reconsider whether a provisional action remains appropriate once prior outcomes, context, and user-specific state are explicitly checked.

This two-pass design provides an explicit verification step before action commitment and is intended to improve the reliability of compact-model decisions. It also supports personalization by explicitly rechecking and updating memory and preference state. (Figure~\ref{fig:two_pass}). At inference time, the host validates the proposed action against safety bounds (Figure~\ref{fig:overview}, right), governs memory persistence through structured update rules that feed back into the personalization layer, and executes actuator updates only when runtime constraints are satisfied.

\begin{figure*}[h!]
\centering
\includegraphics[width=\textwidth]{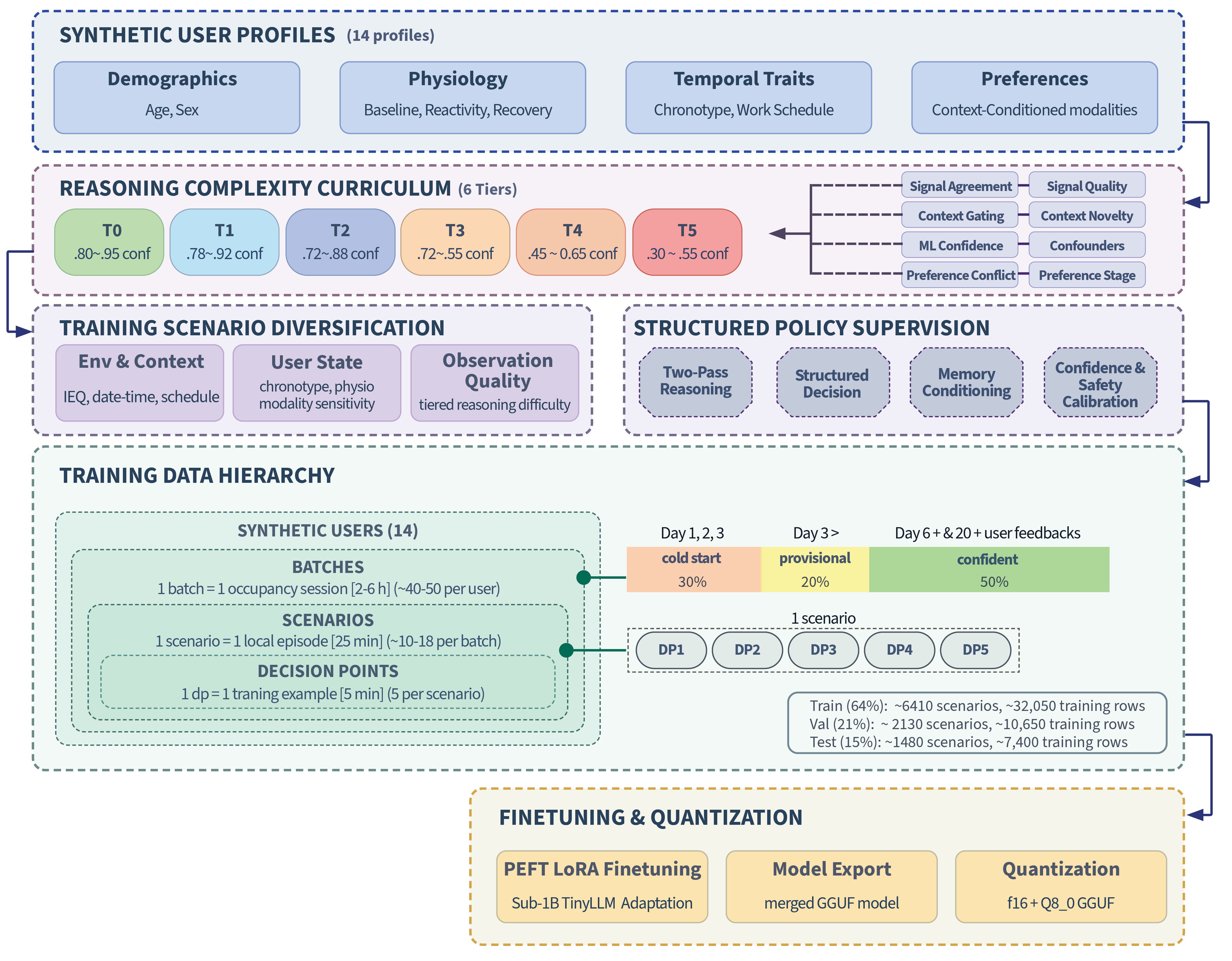}
\caption{\textbf{Affective Agent Training Workflow.} The workflow proceeds from synthetic scenario generation (left), through a six-tier reasoning-complexity curriculum (center), to programmatic two-pass supervision (right).}
\label{fig:workflow}
\end{figure*}

\section{Affective Agent Workflow}
\label{sec:workflow}

Section~\ref{sec:architecture} defines what the architecture must do. This section describes how we instantiate the architecture for evaluation. Realizing this design on resource-constrained edge hardware motivates the use of a compact sub-1B language model, placing this work in the emerging space between TinyML (which has focused on classification and keyword spotting) and full-scale agentic AI (which assumes cloud resources). 

Compared with larger cloud-based models, tiny models have less capacity to internalize specialized domain knowledge, handle conflicting evidence, and infer multi-step decision logic. Rather than scaling the model, the workflow compensates through structured supervision (Figure~\ref{fig:workflow}). Its three building blocks are a chronology-preserving synthetic scenario generation pipeline, a reasoning-complexity curriculum, and programmatic two-pass reasoning.

\subsection{Synthetic Data Generation} 
\label{sec:synthetic_data}

The synthetic training data cohort is not intended to reproduce a specific participant population or empirical data, but to produce training scenarios that confront the model with a range of reasoning challenges: noisy signals, conflicting evidence, uncertainty, and deferral. The evaluation criterion is whether the agent learns to reason over this problem structure; whether those decisions transfer to real-world intervention outcomes is a question for future field study. The synthetic data provide controlled longitudinal variation in physiology, signal quality, and context. This allows the sub-1B models to encounter repeated decisions with noisy and conflicting evidence rather than independent, idealized examples. This evaluates whether a compact model can combine these factors when deciding whether, when, and how to intervene.

The generator is built around a persistent synthetic cohort of 14 users aged 20 to 30, with physiological priors informed by WESAD for stress, SWELL-KW for cognitive load, and En-Gage for thermal comfort~\cite{Schmidt2018WESAD,Koldijk2014SWELLKW,Gao2022EnGage}. The 14-user cohort is designed to represent meaningful inter-user diversity in baseline physiology, reactivity, chronotype, and preference distribution across the training space, rather than to serve as a clinical sample. WESAD (N=15) and SWELL-KW (N=25) inform the physiological priors for these profiles. (Figure~\ref{fig:workflow}, top). 

\subsubsection{Data Hierarchy}
Because deployed decisions depend not only on current evidence but also on what the system has previously observed about a user, the training data preserve the chronological structure of real decision-making (Figure~\ref{fig:workflow}). At the highest level, a batch represents one continuing occupancy session, during which memory and environmental carry-over persist. Within a batch, a scenario is a 25-minute local episode comprising five consecutive 5-minute decision windows. This structure trains the model to reason under partial observation early in an episode and to update its judgment as evidence accumulates.

\subsubsection{Synthetic Data Validation}

\begin{figure}[!h]
\centering
\includegraphics[width=\columnwidth]{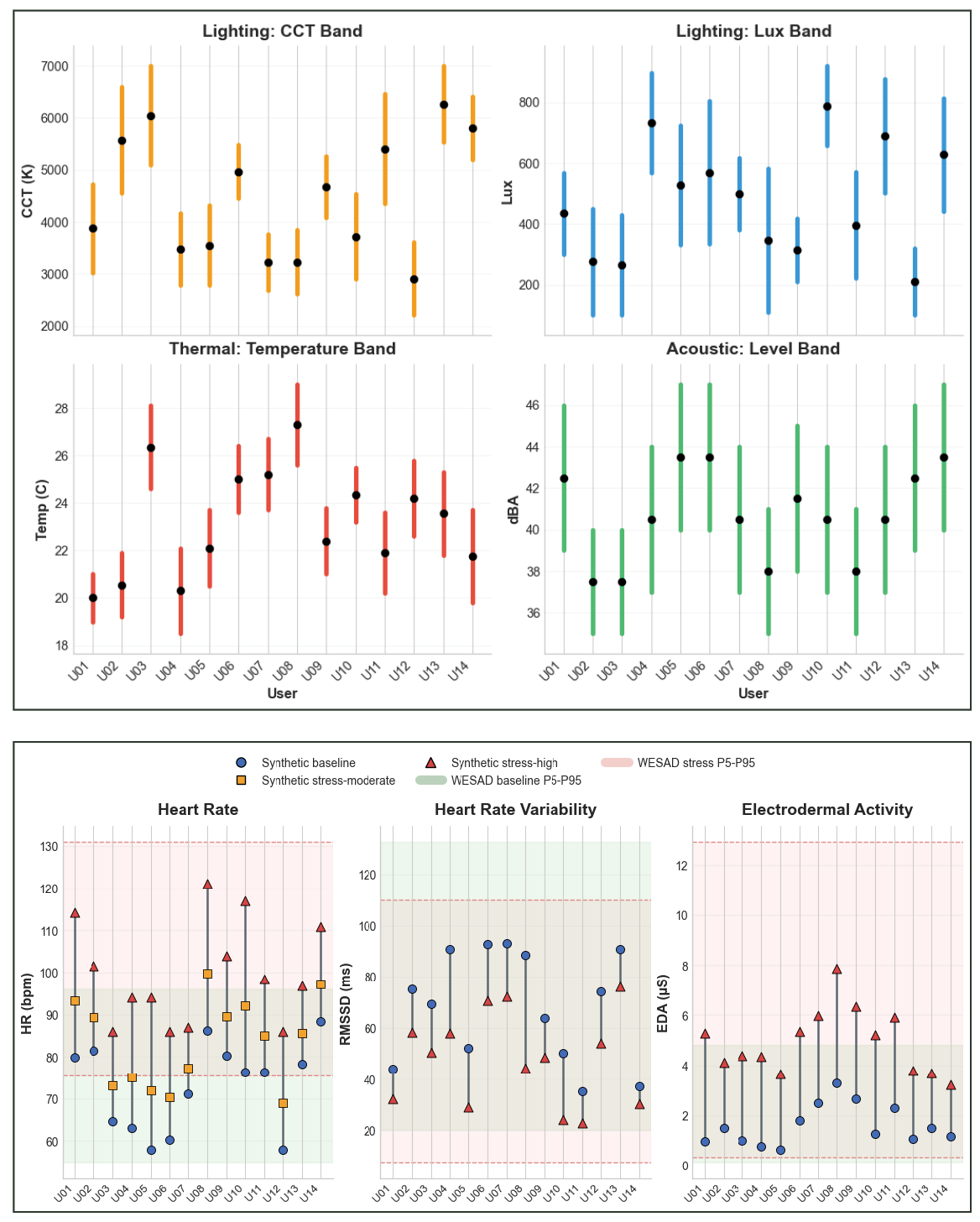}
\caption{\textbf{Synthetic Data Validation.} (Top) Per-user environmental condition distributions across the synthetic cohort, diversity of intervention contexts. (Bottom) Synthetic physiological signal distributions against WESAD bounds, showing the relationship between synthetic physiological distributions and WESAD-derived bounds.}
\label{fig:realism}
\end{figure}

Generating physiological signals for training requires balancing realism against the risk of overfitting to any single laboratory protocol. The synthetic physiology pipeline, therefore, targets field-plausible wearable realism rather than exact reproduction of a specific dataset (Figure~\ref{fig:realism}, Bottom). We generate signals mechanistically using per-user reactivity parameters informed by prior physiological datasets and models ~\cite{Hongn2025StressExercise, Takahashi2021JOS3}. To reflect real wearable use, we inject realistic artifacts, including sensor dropout, intermittent loss, clipping, poor contact, and motion artifacts, to train the model under noisy conditions representative of ambulatory wearable use. 

Ground truth is a bounded supervisory intervention target derived from the simulator's latent state, not a physiological label or self-report. Each target is determined by signal quality, preference life-cycle stage, prior intervention outcomes, and safety constraints that the agent cannot directly observe, and is specified as an acceptable range rather than an exact scalar. Source-side audits verify internal consistency between assigned targets and the state that produced them. This controlled setting supports direct evaluation of decision quality, component ablations, and failure cases across noisy, context-dependent, longitudinal scenarios. It establishes the computational feasibility of the reasoning architecture, while transfer across the broader behavioral variation of real users remains a question for field study.

\subsection{Reasoning Complexity Curriculum}
\label{sec:curriculum}
To expose the model to the full range of intervention-reasoning difficulty, training examples are organized into six complexity tiers, T0 through T5 (Table~\ref{tab:tiers}), designed to teach both confident action selection under clear evidence and appropriate uncertainty calibration under ambiguity. Each tier is defined by the interaction between signal clarity, preference life-cycle maturity, contextual constraints, and prior outcome history. A quota-managed budget maintains proportional coverage, so the model is not dominated by the most frequent cases under unconstrained generation. 

For each scenario, the two-pass reasoning chain is generated programmatically from ground-truth scenario conditions, preference history, and intervention outcomes, rather than distilled from a larger model or produced via an unconstrained chain-of-thought. Because each reasoning step is derived deterministically from known simulator state, the supervision remains internally consistent and avoiding unconstrained model-generated intermediate reasoning.

\begin{table}[t]
\caption{\textbf{Reasoning Complexity Curriculum:} Six tiers (T0--T5) organized by reasoning difficulty and expected confidence range.}
\label{tab:tiers}
\centering
\footnotesize
\renewcommand{\arraystretch}{1.15}
\begin{tabular*}{\columnwidth}{@{\extracolsep{\fill}} p{0.06\columnwidth} p{0.21\columnwidth} p{0.43\columnwidth} p{0.13\columnwidth}}
\toprule
\textbf{Tier} & \textbf{Name} & \textbf{When It Applies} & \textbf{Conf.} \\
\midrule
T0 & Population Default & Clear signals, no confirmed preferences & .80--.95 \\
T1 & Context-Gated & Context blocks action despite plausible intervention evidence & .78--.92 \\
T2 & Learned Pref & Confirmed preferences with clear or moderate signals and no conflict & .72--.88 \\
T3 & Conflicting & Confirmed preferences with conflict or ambiguity & .55--.75 \\
T4 & Provisional & Provisional or cold-start user with unclear signals & .45--.65 \\
T5 & Deep Reasoning & Novel or genuinely ambiguous case with multi-factor uncertainty & .30--.55 \\
\bottomrule
\end{tabular*}
\end{table}

\section{Evaluation}
\label{sec:evaluation} 

Section~\ref{sec:intro} defined Affective Agent through five design properties. This section tests whether the architecture satisfies them within a controlled synthetic evaluation. Because the contribution is architectural, we evaluate decision quality on held-out simulator-generated longitudinal scenarios and deployment feasibility on an embedded edge platform. The scenarios provide controlled variation in signal quality, context, intervention history, covariates, and preference state. The evaluation tests the feasibility of the reasoning substrate rather than real-world intervention efficacy.

\subsection{Experimental Setup}

Evaluating whether memory improves decisions, or whether two-pass reasoning outperforms single-pass reasoning requires knowing a consistent simulator-derived reference decision for each scenario. We evaluate Qwen2.5-0.5B and Qwen3-0.6B, both fine-tuned with the same LoRA configuration and exported in Q8\_0 GGUF format. The deployment feasibility of Affective Agent's on-device reasoning pipeline is benchmarked on a Raspberry Pi Compute Module 5 with no data leaving the device. Reported latency covers prompt-to-output model inference and excludes sensing, feature extraction, classifier execution, prompt assembly, and host-side validation; model loading is measured separately. As a non-learning baseline, we use a deterministic rule-based controller, consistent with evidence that such controllers remain common in deployed control settings~\cite{Xu2026PCMReview}. Two ablations on Qwen3-0.6B isolate component contributions: \emph{no-memory} masks all prompt-visible memory inputs while preserving two-pass reasoning, and \emph{one-pass} preserves memory but removes the verification pass.

We measure five metrics: \emph{action accuracy} and \emph{macro-F1} for correctness on the three-way decision (intervene, wait, abstain); \emph{modality accuracy} for correct intervention channel selection; \emph{schema compliance rate} for structured contract validity; and \emph{ECE} for alignment between reported confidence and actual accuracy. All configurations are evaluated on the same 1{,}600 decision points. Each uses one fine-tuned checkpoint with greedy decoding. QoS uncertainty is estimated by cluster bootstrap over longitudinal scenarios, preserving within-scenario dependence.

\subsection{Results}

\begin{table}[!h]
\caption{\textbf{Quality-of-Service \& Edge Efficiency Results:}
(a)~Main QoS comparison, (b)~ablations on Qwen3-0.6B,
(c)~edge efficiency on Raspberry Pi CM5. \textit{All evaluated on the same 1{,}600 decision points. For QoS, $\pm$ denotes cluster-bootstrap standard error. Latency reported as mean $\pm$ SD. Prefill/decode estimated by token-time regression ($R^2{>}0.999$)}}
\label{tab:qos_combined}
\centering
\scriptsize
\renewcommand{\arraystretch}{1.05}
\setlength{\tabcolsep}{1.5pt}

\begin{tabular}{@{}p{0.33\columnwidth} c c c@{}}
\toprule
\multicolumn{4}{l}{\textbf{(a) Main QoS comparison}} \\
\textbf{Metric}
& \textbf{Qwen3}
& \textbf{Qwen2.5}
& \textbf{Baseline} \\
\midrule
Action Accuracy
& 0.782\,{\scriptsize$\pm$.042}
& 0.811\,{\scriptsize$\pm$.044}
& 0.482\,{\scriptsize$\pm$.037} \\
Macro-F1
& 0.616\,{\scriptsize$\pm$.031}
& 0.677\,{\scriptsize$\pm$.085}
& 0.431\,{\scriptsize$\pm$.029} \\
Modality Accuracy
& 0.864\,{\scriptsize$\pm$.039}
& 0.701\,{\scriptsize$\pm$.075}
& 0.745\,{\scriptsize$\pm$.184} \\
Schema Compliance
& 0.993
& 0.993
& 1.000 \\
ECE
& 0.301\,{\scriptsize$\pm$.024}
& 0.255\,{\scriptsize$\pm$.029}
& 0.097\,{\scriptsize$\pm$.045} \\
\midrule
\multicolumn{4}{l}{\textbf{(b) Ablations}} \\
\textbf{Metric}
& \textbf{Full}
& \textbf{No-Memory}
& \textbf{One-Pass} \\
\midrule
Action Accuracy
& 0.782\,{\scriptsize$\pm$.042}
& 0.731\,{\scriptsize$\pm$.031}
& 0.715\,{\scriptsize$\pm$.048} \\
Macro-F1
& 0.616\,{\scriptsize$\pm$.031}
& 0.501\,{\scriptsize$\pm$.035}
& 0.463\,{\scriptsize$\pm$.034} \\
Modality Accuracy
& 0.864\,{\scriptsize$\pm$.039}
& 0.605\,{\scriptsize$\pm$.082}
& 0.652\,{\scriptsize$\pm$.078} \\
Schema Compliance
& 0.993
& 1.000
& 1.000 \\
ECE
& 0.301\,{\scriptsize$\pm$.024}
& 0.321\,{\scriptsize$\pm$.048}
& 0.377\,{\scriptsize$\pm$.049} \\
\midrule
\multicolumn{4}{l}{\textbf{(c) Edge efficiency}}\\
\textbf{Metric}
& \textbf{Qwen3}
& \textbf{Qwen2.5}
& \\
\midrule
Total Latency (s)
& $83.358 \pm 5.428$
& $53.657 \pm 3.211$ 
& -- \\
\quad Prefill Time (s)
& 20.125
& 8.204
& -- \\
\quad Decode Time (s)
& 63.233
& 45.453
& -- \\
Decode Rate (tok/s)
& 14.1
& 19.7
& -- \\
Completion Tokens
& 892.125
& 897.508
& -- \\
\midrule
Load Time (s)
& 0.447
& 1.212
& -- \\
GGUF Size (MB)
& 609.8
& 506.5
& -- \\
Peak RSS (MB)
& 1861.59
& 2277.77
& -- \\
RAM Util. (\%)
& 23.1
& 28.3
& -- \\
\midrule
Mean SoC Temp.\ ($^\circ$C)
& 62.0
& 57.5
& -- \\
Peak SoC Temp.\ ($^\circ$C)
& 63.9
& 59.5
& -- \\
\bottomrule
\end{tabular}
\end{table}

\subsubsection{Main QoS Results}
\textbf{Learned reasoning outperforms the rule baseline.} Qwen3 and
Qwen2.5 achieve action accuracies of 0.782 and 0.811, compared with 0.482 for the rule-based baseline, corresponding to improvements of 30.0 and 32.9 percentage points. These results show that the learned models handle the context, history, and uncertainty represented in the evaluation scenarios more effectively than fixed threshold logic.

\textbf{Performance tradeoff across the two agents.} A metric inversion emerges between the two models. Qwen2.5-0.5B achieves higher top-level action accuracy (0.811 vs.\ 0.782) and macro-F1 (0.677 vs.\ 0.616), while Qwen3-0.6B achieves higher modality accuracy (0.864 vs.\ 0.701). The pattern reflects checkpoint-specific differences between top-level intervention decisions and modality routing, rather than evidence that one backbone specializes in deciding \emph{whether} to act and the other in deciding \emph{how}. That both instantiations exceed the baseline on action accuracy and macro-F1 indicates that Affective Agent is not tied to a single fine-tuned checkpoint.

\textbf{Decoding dominates edge latency.} Decoding the approximately 900-token two-pass completion accounts for 75.9\% of mean latency for Qwen 3 and 84.7\% for Qwen 2.5. Mean latency occupies 27.8\% and 17.9\% of the five-minute total sensing interval, respectively. Because inference occupies a meaningful portion of each sensing interval, future deployments should revalidate user context and safety constraints immediately before actuation and discard decisions that have become stale.

\textbf{Schema compliance is near-perfect } Both instantiations achieve 0.993 compliance with the structured decision contract, indicating reliable schema-constrained output generation at sub-1B scale. 

\textbf{Calibration remains the QoS limitation.} The agents’ ECE (0.255--0.301) exceeds that of the rule-based baseline (0.097), indicating poorer alignment between reported confidence and observed accuracy despite stronger intervention-decision performance. The host mitigates this limitation by enforcing action bounds and requiring abstention below a confidence threshold, rather than relying on the model's self-reported confidence alone. Future work should evaluate post-hoc calibration and calibration-aware training.

\subsubsection{Reasoning and Qualitative Analysis}
\textbf{Two-pass reasoning produces structured assessment and verification.} The two-pass structure often produces qualitatively coherent assessment and verification outputs (Figure~\ref{fig:two_pass}). The first pass organizes physiological evidence, signal quality, context, and prior outcomes into a provisional state interpretation, while the second pass re-checks that judgment against recent history, preference maturity, contextual permissibility, and confidence.

\textbf{Verification contributes beyond the initial assessment.} In the examined outputs, the second pass refines modality-specific reasoning, tempers personalization when evidence is weak, and revises confidence when the initial assessment conflicts with context or history. The ablation supports this interpretation. Removing the second pass reduces macro-F1 from 0.616 to 0.463 and increases ECE from 0.301 to 0.377 (Table~\ref{tab:qos_combined}), supporting the contribution of verification to the evaluated decision and confidence-calibration metrics. These examples illustrate the structure of the generated reasoning but do not independently establish its faithfulness.

\subsubsection{Ablation Results}
The ablations isolate the contribution of two novel components (memory-driven personalization and two-pass reasoning) to the design properties (Table~\ref{tab:qos_combined}).

\textbf{Memory matters most for personalized routing.} Removing memory (property~2) reduces macro-F1 from 0.616 to 0.501 and modality accuracy from 0.864 to 0.605. The modality accuracy drop is especially large, indicating that memory conditioning primarily enables correct intervention routing rather than coarse action selection. This also validates property~(3): personalization depends on host-managed evolving state rather than on user-specific weight updates.

\textbf{Two-pass reasoning is essential for reliability.} Removing the verification pass (property~4) degrades macro-F1 to 0.463 and worsens ECE from 0.301 to 0.377. Without explicit verification, the model loses not only decision quality but also the structured reasoning that makes its outputs interpretable and auditable.

\section{Discussion}
\label{sec:discussion}

Three findings emerge from the synthetic evaluation. First, sub-1B models can execute structured intervention reasoning on-device, not just classification. Second, host-managed memory improves personalized intervention without per-user retraining. Third, an explicit two-pass reasoning chain meaningfully improves both reliability and confidence calibration in compact models. 

These results suggest a practical path for wearable intervention systems. Instead of scaling models or using the cloud, limited capacity can be offset through structured memory, supervised reasoning, and host-managed state.

The architecture is domain-agnostic. While we evaluate on IEQ as one exemplar, the same three-layer design may apply wherever a wearable system must decide whether, when, and how to intervene. The current simulator does not capture mood-driven rejection, situational preference shifts, or competing task and social demands.

\section{Conclusion}
\label{sec:conclusion}

Affective Agent advances wearable computing from passive state inference toward structured, personalized intervention reasoning on-device. Within the synthetic evaluation, episodic and semantic memory improve personalized intervention routing without per-user retraining, while two-pass reasoning improves decision quality and confidence calibration in compact models.

The architecture and methodology contribute to several intersecting areas. For edge AI and TinyML, the results show that sub-1B models can support structured decision-making beyond classification, opening new task categories for on-device intelligence. For wearable system designers, the architecture offers a reference blueprint for closing the loop from sensing to personalized intervention across application domains, including affective computing, digital health, and smart home systems. The reasoning-complexity curriculum and evaluation methodology may also inform future benchmarking efforts for tiny agentic systems.

Future work should evaluate the architecture longitudinally with real users, improve confidence calibration and decoding latency, measure power directly, and investigate more aggressive quantization and NPU acceleration for wrist-worn deployment.

\def\refname{REFERENCES}
\vspace*{-8pt}


\begin{IEEEbiography}{Reina Mun}{\,} is a doctoral candidate in computational design and HCI at Harvard University. Her research focuses on multimodal and adaptive systems for ambient intelligence, with applications in wearable health and intelligent environments. Contact her at reina\_mun@gsd.harvard.edu. \vspace*{8pt}
\end{IEEEbiography}

\begin{IEEEbiography}{Zishen Wan}{\,} is the Postdoctoral Fellow in Electrical Engineering at Harvard University. His research focuses on computer architecture, with an emphasis on cross-stack co-design of systems, architectures, and silicon for physical intelligence. He holds a Ph.D. in Electrical and Computer Engineering from Georgia Institute of Technology. Contact him at zishenwan@seas.harvard.edu. \vspace*{8pt}
\end{IEEEbiography}

\begin{IEEEbiography}{Vijay Janapa Reddi} {\,} is the Gordon McKay Professor of Electrical Engineering at Harvard University. His research focuses on the intersection of computer architecture, machine learning systems, and autonomous agents, spanning mobile, edge, and IoT platforms. He is Vice President and co-founder of MLCommons and serves on the board of the EDGE AI Foundation. He holds a Ph.D.\ in computer science from Harvard University. Contact him at vj@eecs.harvard.edu. \vspace*{8pt}
\end{IEEEbiography}

\end{document}